\documentclass[letterpaper, 10 pt, conference]{ieeeconf}  

\IEEEoverridecommandlockouts                              

\usepackage{times}
\usepackage{amssymb}
\usepackage{amsmath}
\usepackage{graphics}
\usepackage{epsfig}
\usepackage{multirow}
\usepackage{tabularx}
\usepackage{nicefrac}
\usepackage{cases}          
\usepackage{subcaption}
\usepackage{mathtools}
\usepackage{xcolor}
\usepackage{algorithm}
\usepackage{algorithmicx}
\usepackage{algpseudocode}
\usepackage{balance}
\usepackage{gensymb}
\usepackage{hyperref}

\usepackage{soul}
\usepackage{adjustbox}

\hypersetup{colorlinks=true}
\hypersetup{linkcolor=red}
\hypersetup{citecolor=blue}
\hypersetup{filecolor=cyan}
\hypersetup{urlcolor=magenta}

\newcommand{\tn}[1]{\textnormal{#1}}
\newcommand{\tb}[1]{\textbf{#1}}

\newcommand{\mat}[0]{\begin{bmatrix}}
\newcommand{\mate}[0]{\end{bmatrix}}

\newcommand{\vd}{\mathbf{d}}

\newcommand{\vf}{\mathbf{f}}

\newcommand{\vh}{\mathbf{h}}

\newcommand{\vn}{\mathbf{n}}
\newcommand{\vp}{\mathbf{p}}

\newcommand{\vu}{\mathbf{u}}
\newcommand{\vv}{\mathbf{v}}
\newcommand{\vw}{\mathbf{w}}
\newcommand{\vx}{\mathbf{x}}

\newcommand{\cC}{\mathcal{C}}

\newcommand{\cF}{\mathcal{F}}

\newcommand{\cO}{\mathcal{O}}
\newcommand{\cP}{\mathcal{P}}

\newcommand{\cR}{\mathcal{R}}
\newcommand{\cS}{\mathcal{S}}

\newcommand{\cU}{\mathcal{U}}
\newcommand{\cV}{\mathcal{V}}
\newcommand{\cW}{\mathcal{W}}

\newcommand{\R}{\mathbb{R}}
\newcommand{\N}{\mathbb{N}}

\newcommand\norm[1]{\left\|#1\right\|}              
\newcommand\abs[1]{\left|#1\right|}                 

\newcommand{\rebuttal}[1]{{\color{blue}#1}}

\title{\LARGE \bf
Unwieldy Object Delivery with Nonholonomic Mobile Base: \\ A Stable Pushing Approach}
\author{Yujie Tang, Hai Zhu, Susan Potters, Martijn Wisse and Wei Pan
\thanks{The authors are with the Department of Cognitive Robotics, Delft University of Technology, 2628
CD, Delft, The Netherlands {\tt\small $\{$y.tang-6, h.zhu, s.potters, m.wisse, wei.pan$\}$@tudelft.nl}}%
}
\begin{document}

\maketitle              

\begin{abstract}

This paper addresses the problem of pushing manipulation with nonholonomic mobile robots. Pushing is a fundamental skill that enables robots to move unwieldy objects that cannot be grasped. We propose a stable pushing method that maintains stiff contact between the robot and the object to avoid consuming repositioning actions. We prove that a line contact, rather than a single point contact, is necessary for nonholonomic robots to achieve stable pushing. We also show that the stable pushing constraint and the nonholonomic constraint of the robot can be simplified as a concise linear motion constraint. Then the pushing planning problem can be formulated as a constrained optimization problem using nonlinear model predictive control (NMPC). According to the experiments, our NMPC-based planner outperforms a reactive pushing strategy in terms of efficiency, reducing the robot's traveled distance by 23.8\% and time by 77.4\%. Furthermore, our method requires four fewer hyperparameters and decision variables than the Linear Time-Varying (LTV) MPC approach, making it easier to implement. Real-world experiments are carried out to validate the proposed method with two differential-drive robots, Husky and Boxer, under different friction conditions.
   
\end{abstract}

\section{Introduction}\label{sec:intro}

\begin{figure}
     \centering
     \begin{subfigure}[b]{0.23\textwidth}
         \centering
         \includegraphics[width=\textwidth, height=0.1\textheight]{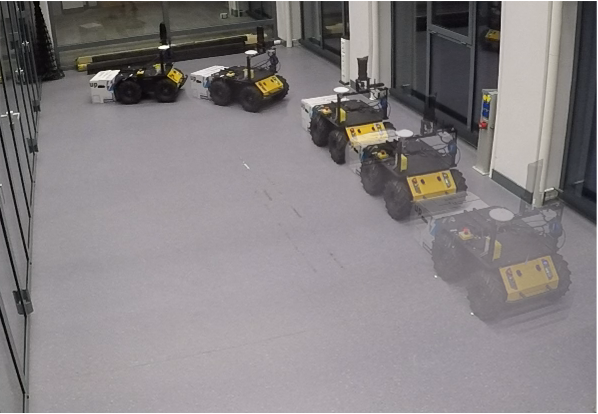}
         \caption{}
     \end{subfigure}
     ~
     \begin{subfigure}[b]{0.23\textwidth}
         \centering
         \includegraphics[width=\textwidth, height=0.1\textheight]{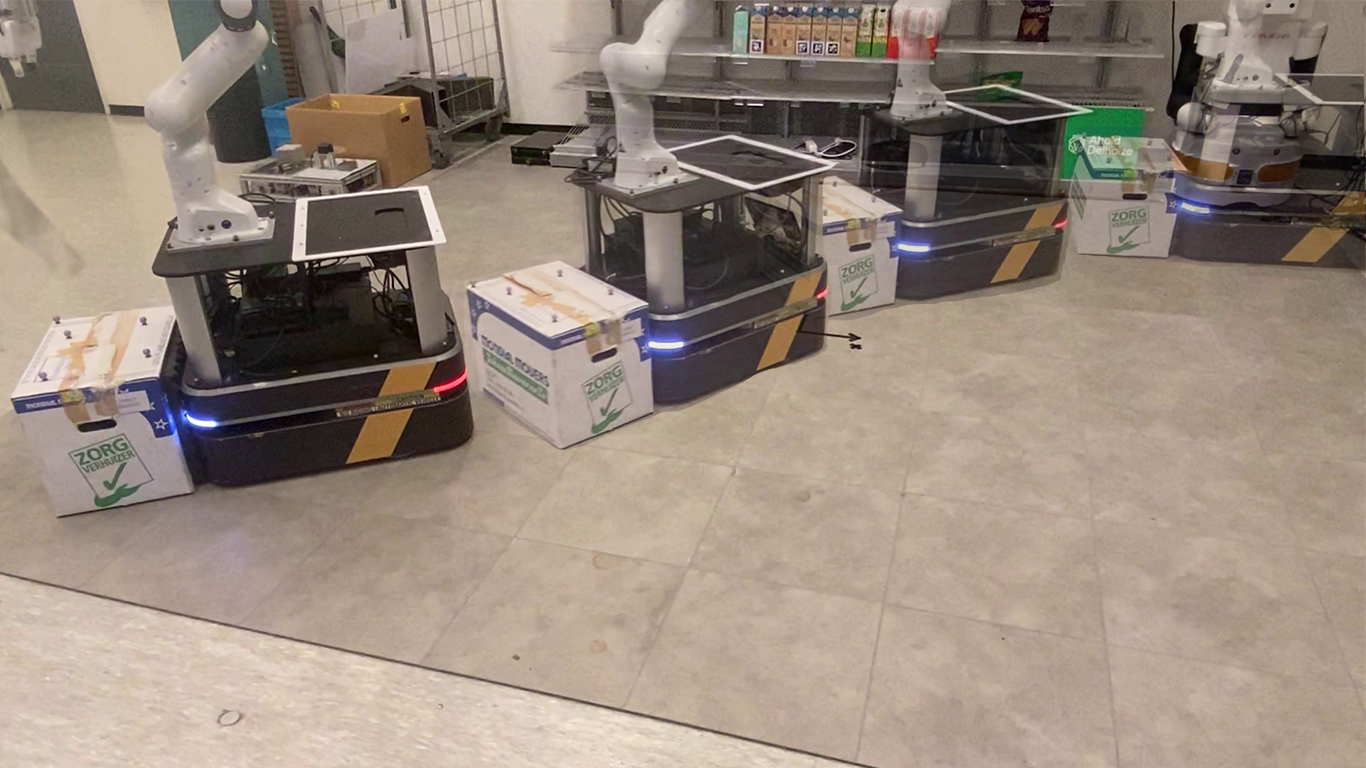}
         \caption{}
     \end{subfigure}
	\caption{
The wheeled mobile robots (Clearpath Husky and Boxer) push a paper box to a goal location and to track a reference path, respectively. Transparency of the robots and box indicates their movement. \label{fig:push}}
 \vspace{-0.6cm}
\end{figure}

With mobile robots increasingly being used, 
there are various scenarios in which the robots are expected to perform additional 
delivery tasks while maneuvering, for example, a robot conveying a package in 
a warehouse. In this regard, mobile robots equipped with robot arms have become 
progressively popular. However, the delivered object may be sometimes unwieldy, either too heavy 
or too large, for the robot arm to grasp. In this case, one option is to 
manipulate the object by pushing it with the robot arm \cite{heins2021mobile}. 
Alternatively, the robot can push the object, as shown 
in Fig. \ref{fig:push}. Without a robot arm, pushing with the robot 
expands its manipulation repertoire, making it not just a mobile base. 
Moreover, it reduces the cost, space, and payload by eliminating 
the robot arm \cite{stuber2020let}.

Research on pushing with mobile robots is still limited, though pushing with robot arms has been extensively studied \cite{chavan2018stable, hogan2020reactive, kopicki2017learning}. Mobile robots have nonholonomic constraints that restrict their ability to freely reach various planned contact points. As a result, the pushed object is prone to sliding away, requiring time-consuming and effort-consuming repositioning actions to restart pushing.
To address this challenge, \cite{lynch1996stable} proposed stable pushing, which involves maintaining a stiff robot-object contact to prevent frequent repositions. This approach can reduce the risk of losing control over the object resulting in improved efficiency.

As concluded in \cite{zhou2019pushing}, stable pushing with a 
single-point contact can be reducible to the Dubins car problem, where the sticking contact constraint is translated to bounded curvatures of the object's trajectory, represented as a motion cone for the object. However, we extend this conclusion by proving that stable pushing is not achievable for a differential-drive mobile robot pushing with a single-point contact, due to the limited friction cone and the nonholonomic constraint of the robot. It can not provide enough friction force to maintain a stiff robot-object contact. As a follow-up study to \cite{zhou2019pushing}, we introduce a line contact to make stable pushing possible where a larger friction cone can be provided. Based on it, we prove that the stable pushing constraint and robot nonholonomic constraint can be combined as a linear motion constraint on the robot's control input, which greatly simplifies the pushing planning problem compared to \cite{bertoncelli2020linear}, as the stable pushing can be guaranteed implicitly with the control constraint.

We formulate the goal-conditioned stable pushing problem as a constrained optimization problem by employing Nonlinear Model Predictive Control (NMPC). 
Our NMPC planner with the concluded motion constraint guarantees that the object's motion is within the motion cone for stable pushing and the physical limitation of the robot is met. The main contributions of this paper can be summarized as follows:
\begin{itemize}
\item We first propose a stable pushing approach for nonholonomic mobile robots that maintains a stiff robot-object contact so that the need for frequent repositioning actions can be minimized.
\item  We then derive a concise linear motion constraint to simplify the stable pushing one in \cite{bertoncelli2020linear} and develop an algorithm that is easier to be implemented with commercial solvers.
\item Lastly, we evaluate the proposed method through real-world experiments using wheeled mobile robots (Clearpath Husky and
Boxer) that showed significant reductions in traveled distance and time.
\end{itemize}

\section{Related Work}\label{sec:relatedWork}
In the class of non-prehensile manipulation, pushing received the most attention for its high flexibility and efficiency in completing a task \cite{emeli2020joint, 1013403, 1545603}. Early research on mobile robot pushing involved using compliance to push the object along the environment boundaries \cite{1545603}. Instead of finding feasible paths in free space, compliance pushing simplifies the problem and provides additional options for finding a pushing path. However, the method is limited to disk-shaped pushers and objects, and can only be applied in environments with smooth boundaries which is rare in the real world. 

In order to achieve practical mobile robot pushing, a reactive pushing controller is proposed in \cite{krivic2019pushing}, where the basic idea is to keep the robot, the object, and the goal in a line so as to push the object toward the goal. Nevertheless, the method is limited to pushing small-sized objects with circular or point-sized robots such that it is easy to reposition around the object to change the pushing direction. Instead of pushing reactively, \cite{mericcli2015push} presents a rapidly-exploring random tree (RRT) based planner that uses past pushing experiences to construct achievable and collision-free pushing plans. However, both \cite{krivic2019pushing} and \cite{mericcli2015push} assume the use of omnidirectional mobile robots, which can freely move around the object to achieve the planned pushing actions. For widely-used differential drive robots, limited research has been conducted, as the nonholonomic constraint hinders their ability to smoothly push around the object, making pushing planning more complex.
 
In addition to control and planning, a significant challenge in mobile robot pushing is the uncertainty about the object’s pose after each action \cite{agboh2020pushing}. The methods discussed above rely on reactive actions taken after observing the resulting motion of the pushed object. The robot pusher and the object strive to maintain an equilibrium configuration to continue moving together, resembling a ``catching'' action during navigation \cite{lynch1996stable}. Thus, the crucial aspect of designing a push/navigation controller is ensuring the stability of this ``catching" action.

The concept of stable pushing, which establishes a predictable stiff contact between the robot and the object, was proposed based on the mechanics of planar sliding in \cite{doi:10.1177/027836498600500303}. This idea has been widely used in the field of pushing manipulation, as demonstrated in \cite{hogan2016feedback, hogan2018reactive}. In this paper, we also adopt the concept of stable pushing and propose a method that enables a differential-drive robot to push an object without losing contact. 
The most related methodology is proposed in \cite{bertoncelli2020linear} where a Linear Time-Varying (LTV) MPC is used for mobile robots to push an object along a given path, where stable pushing is achieved by optimizing for both the pushing force and the robot control inputs, which explicitly imposes the friction cone constraints. However, we found that it is computationally expensive to solve this optimization problem due to the additional decision variables and constraints. Furthermore, it is not even solvable with commercial solvers such as ACADOS \cite{Verschueren2021}. To address it, a reference trajectory and supplementary linearization are essential components in the solution process. In contrast, our proposed method implicitly constrains the stiff robot-object contact by deriving a concise motion constraint for the robot control input, making it easier to implement. The validation of the proposed method is also shown in both simulation and real-world experiments.

\section{Preliminaries}\label{sec:preliminary}

Throughout this paper, scalars are denoted by italic lowercase letters, e.g., $x$, vectors 
by bold lowercase, e.g., $\vx$, matrices by plain uppercase, e.g., $A$, and sets by calligraphic uppercase, e.g., $\cC$. The superscript $\vx^\top$ or $A^\top$ denotes the 
transpose of a vector $\vx$ or a matrix $A$.  Denote by $\{\cW\}$, $\{\cR\}$, and $\{\cO\}$, 
the global world frame, the robot body frame, and the object body frame, respectively. 

\subsection{Robot dynamics model}
Consider a nonholonomic differential-drive robot. Let $\vx_{\tn{r}}=[x_{\tn{r}}, y_{\tn{r}}, \theta_{\tn{r}}, v_{\tn{r}}, \omega_{\tn{r}}]^\top \in \R^5$ denote the robot state vector, where $\vp_{\tn{r}} = [x_{\tn{r}}, y_{\tn{r}}]^\top$ represents the robot position in the world frame $\{\cW\}$, $\theta_{\tn{r}}$ its orientation and $v_{\tn{r}}$ and $\omega_{\tn{r}}$ its linear and angular velocities referring to the world frame, as shown in Fig. \ref{fig:frame}. Denote by $\vu_{\tn{r}} = [a_{\tn{r}}, \xi_{\tn{r}}]^\top \in \R^2$ the robot's control input vector, in which $a_{\tn{r}}$ and $\xi_{\tn{r}}$ are its linear and angular accelerations, respectively. The robot dynamics are described by the following nonlinear differential equations \cite{tzafestas2013introduction}:
\begin{equation}
	\begin{bmatrix}
		\dot{x}_\tn{r}\\ \dot{y}_\tn{r}\\ \dot{\theta}_\tn{r}\\ \dot{v}_\tn{r}\\\dot{\omega}_\tn{r}
	\end{bmatrix}
	=
	\begin{bmatrix}
		{v}_{\tn{r}}\cos\theta_\tn{r} \\
		{v}_{\tn{r}}\sin\theta_\tn{r} \\
		\omega_\tn{r} \\
		0\\
		0
	\end{bmatrix}
	+
	\begin{bmatrix}
		0 \\
		0 \\
		0 \\
		a_\textnormal{r} \\
		\xi_\tn{r}
	\end{bmatrix},
\end{equation}
which can further be written in a nonlinear discrete form $\vx_{\tn{r}}^{t+1} = \vf_{\tn{r}}(\vx_{\tn{r}}^t, \vu_{\tn{r}}^{t})$, where $t \in \N$ denotes the time step. 

The robot velocity expressed in the robot frame is ${}^{ \cR} \vv_\tn{r}= [{v}_{\tn{r}} ,0]^\top$. By transforming it into the world frame, we can achieve 
\begin{equation}
    {}^{\cW} \vv_\tn{r}={}^{\cW} R_{\cR} {}^{ \cR} \vv_\tn{r} =     	\begin{bmatrix}
		\cos \theta_\tn{r} &-\sin \theta_\tn{r} \\
		\sin \theta_\tn{r} &\cos \theta_\tn{r} 
\end{bmatrix} 
\begin{bmatrix}
{v}_{\tn{r}} \\
0
  \end{bmatrix},
\label{eq:robot_velocity_in_world}
\end{equation}
where ${}^{\cW} R_{\cR}$ represents the rotation matrix that transforms from the robot frame, ${\cR}$, to the world frame, $\cW$.

\subsection{Quasi-static pushing}\label{sec:}

Pushed by the mobile robot, the object slides with friction interaction with both the 
ground and the robot. The friction interaction is assumed to conform to Coulomb’s law. A quasi-static assumption is made here that the motion of the system is slow and the wrenches are balanced with negligible inertia effects. Then, a force-motion mapping can be given according to the Limit Surface theory proposed in 
\cite{goyal1991planar}. 
All the possible static and sliding friction wrenches form a convex 
set whose boundary is called limit surface. 
Under the uniform pressure distribution, the limit surface is a closed convex surface
and can be approximated by an ellipsoid \cite{lee1991fixture}. 
In this case, the applied push wrench that 
quasi-statically balances the friction wrench has:
\begin{equation}\label{eq:limit_surface}
        {}^{\cO}\vw_{\tn{p}}^\top H {}^{\cO}\vw_{\tn{p}} = 1,
\end{equation}
in which $H=\tn{diag}(\frac{1}{(\mu_{\tn{g}} N_{\tn{o}})^2},
\frac{1}{(\mu_{\tn{g}} N_{\tn{o}})^2},
\frac{\gamma_{\tn{g}}^2}{(\mu_{\tn{g}} N_{\tn{o}})^2})$,
where ${}^\cO\vw_{\tn{p}} = [{}^\cO f_{\tn{p},x}, {}^\cO f_{\tn{p},y}, 
{}^\cO \tau_{\tn{p}}]^\top \in \R^3$ denotes the wrench applied by the pusher that quasi-statically 
balances the friction wrench exerted by the ground planar surface,
the left super-script ${}^\cO\cdot$ represents variables in the object body frame. 
$\mu_{\tn{g}}$ is the friction coefficient between the object and the ground planar surface, 
$N_{\tn{o}}$ the gravity of the object, 
and $\gamma_{\tn{g}}$ an integration constant related to the contact surface area 
\footnote{$\gamma_{\tn{g}} = \frac{A(\cS_{\tn{g}})}{\iint_{\cS_{\tn{g}}}\sqrt{x^2+y^2}dxdy}$, 
where $\cS_{\tn{g}}$ is the contact patch between the object and the ground planar surface, 
and $A(\cS_{\tn{g}})$ its area.}.

The friction wrench is a point on the limit surface when the object is sliding. Moreover, the direction of the object's twist ${}^\cO\vv_{\tn{o}}= [{}^\cO v_{\tn{o},x}, {}^\cO v_{\tn{o},y}, {}^\cO \omega_{\tn{o}}]^\top \in \R^3$ is given by the normal to the limit surface at that point \cite{goyal1991planar}. Hence, there is: 
\begin{equation}\label{eq:force_motion_model}
    {}^\cO\vv_{\tn{o}} \propto \frac{\partial}{\partial {}^\cO\vw_{\tn{p}}}({}^\cO\vw_{\tn{p}}^\top H {}^\cO\vw_{\tn{p}}) \propto H {}^\cO\vw_{\tn{p}}.
\end{equation}
 
\subsection{Dubins car model with a single-point contact pusher}\label{sec:graphical_derivation_single_point}

\begin{figure}[t]
	\begin{centering}
		\includegraphics[width=0.8\columnwidth]{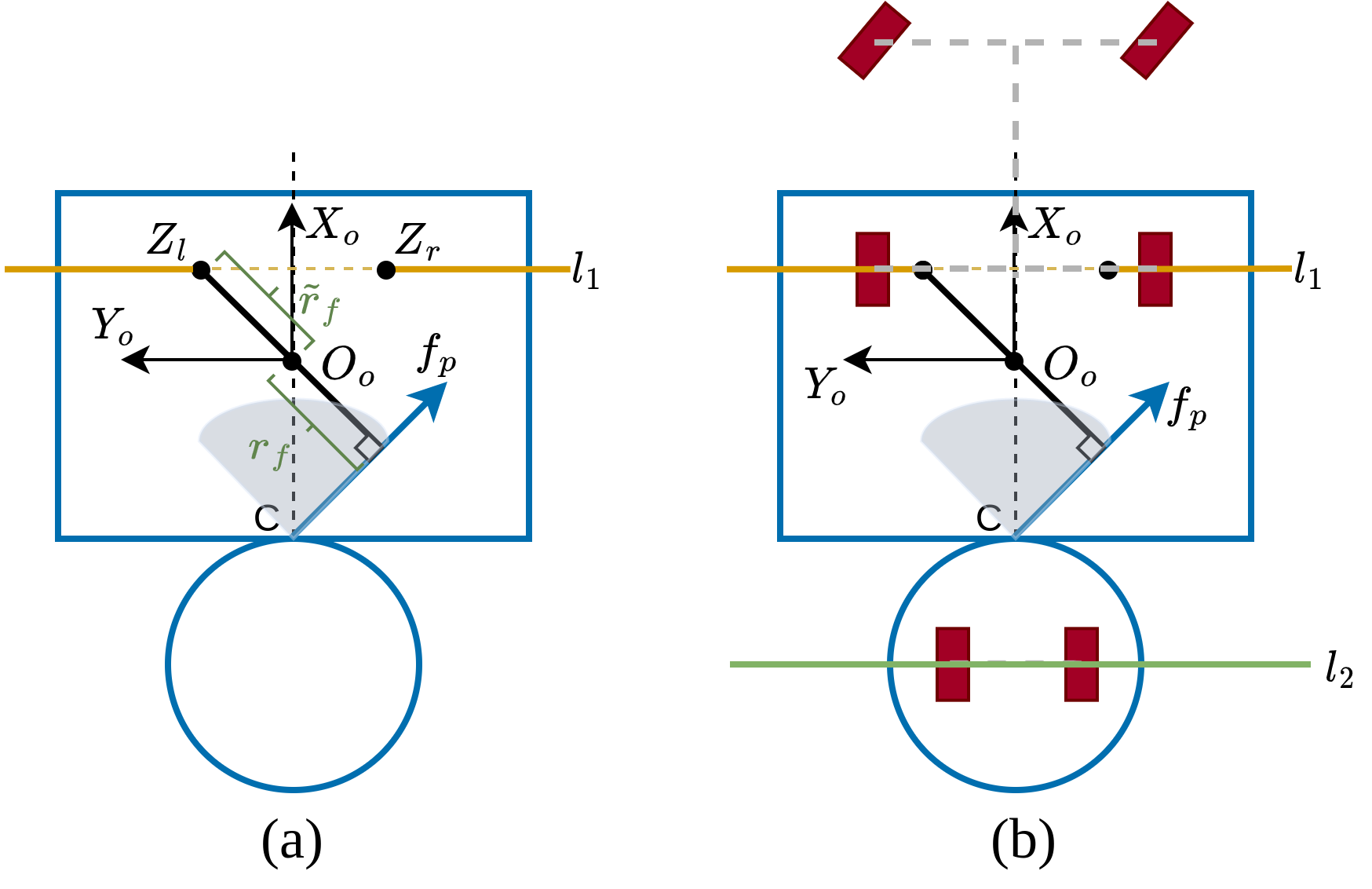}
		\par\end{centering}
	\caption{ Illustration of the possible center of rotation. The circle and the rectangle represent the robot and the object in a 2D plane. The grey area and the blue arrows respectively indicate the friction cone and its edges. The orange line indicates the set of rotation centers of the object under stable push. While the green line is the robot's common left and right wheel axis and the line of its possible rotation centers. In (a), the object is pushed by an omni-directional pusher with a point contact. The set of its possible rotation centers lies on the orange line. In (b), the object is pushed by a nonholonomic robot with its rotation centers on the green line. However, there is no overlap between the possible rotation center of the wheeled robot and the object under this contact configuration. The red bricks, connected by grey dashed lines, represent the wheels of a car model in Dubin's car problem.     \label{fig:cor}
   }
\end{figure}

As concluded in \cite{zhou2019pushing}, stable pushing with a single-point contact can be reducible to the Dubins car problem \cite{dubins1957curves}.
As shown in Fig. \ref{fig:cor}a, a round pusher pushes a rectangle shaped object at point $C$ with a pushing force ${}^\cO\vf_\tn{p} = 
[f_\tn{p,x}, f_\tn{p,y}]$, which is limited within the friction cone. The resulted twist of the object, ${}^\cO \vv_{\tn{o}}$, can be represented
as an instantaneous center of rotation $IRC = [{}^\cO v_{\tn{o},x}/{}^\cO \omega_{\tn{o}}, 
{}^\cO v_{\tn{o},y}/{}^\cO \omega_{\tn{o}}]$.

Given a pushing force ${}^\cO \vf_p$ at contact point $C$, the distance from the object frame origin $O_o$ to the line of force is $
    r_\tn{f}= \frac{\abs{ ^{\cO} x_\tn{c} f_\tn{p,x}}}{\sqrt{f_\tn{p,x}^2 + f_\tn{p,y}^2}}$.
According to the limit surface theory, distance from the center of rotation to the origin is inverse-proportional to $r_\tn{f}$,
that is, $\tilde{r}_\tn{f}=\sqrt{\frac{{}^\cO v_{\tn{o},x}^2 + {}^\cO v_{\tn{o},y}^2}{{}^\cO v_{\tn{o},\omega}^2}} =  \frac{\gamma_{\tn{g}}^2}{ r_\tn{f}}$.

It is demonstrated in \cite{zhou2019pushing}, as in Projective Geometry, the dual of the line of pushing force $\vf_\tn{p}$ about the origin $O_o$ is the instantaneous center of rotation, $IRC$. So the dual of $\vf_p$ in all directions forms a line, as a set of all the possible instantaneous rotation centers, which is perpendicular to the vector from the origin to the contact point, represented as the dashed orange line, $l_1$, in Fig. \ref{fig:cor}a. But due to the friction cone constraint, the rotation center will not be positioned on the line segment $Z_l Z_r$ whose two vertices correspond to the pushing force along the edge of the friction cone.

In other words, the stable pushing constraint is translated 
to a bounded curvature of the object, which makes the stable pushing planning a Dubins car problem, as depicted in Fig. \ref{fig:cor}b.
However, \cite{zhou2019pushing} only considers the omnidirectional pushers. If we take a differential-drive wheeled robot as the pusher, 
the robot can only rotate about a point that lies along its common left and right wheel axis \cite{saidonr2011differential}, as shown in Fig. \ref{fig:cor}b. There comes the contradiction that the shared rotation center of the robot and the object can only be the intersection of $l_1$ and $l_2$, which means the robot and
the object can only move together straightly forward or rotate around the intersection point of the two lines of rotation centers to maintain stable pushing. 

\section{Sticking contact constraint} \label{sec:method}

\begin{figure}[t]
\vspace{-5mm}
	\begin{centering}
		\includegraphics[width=0.7\columnwidth]{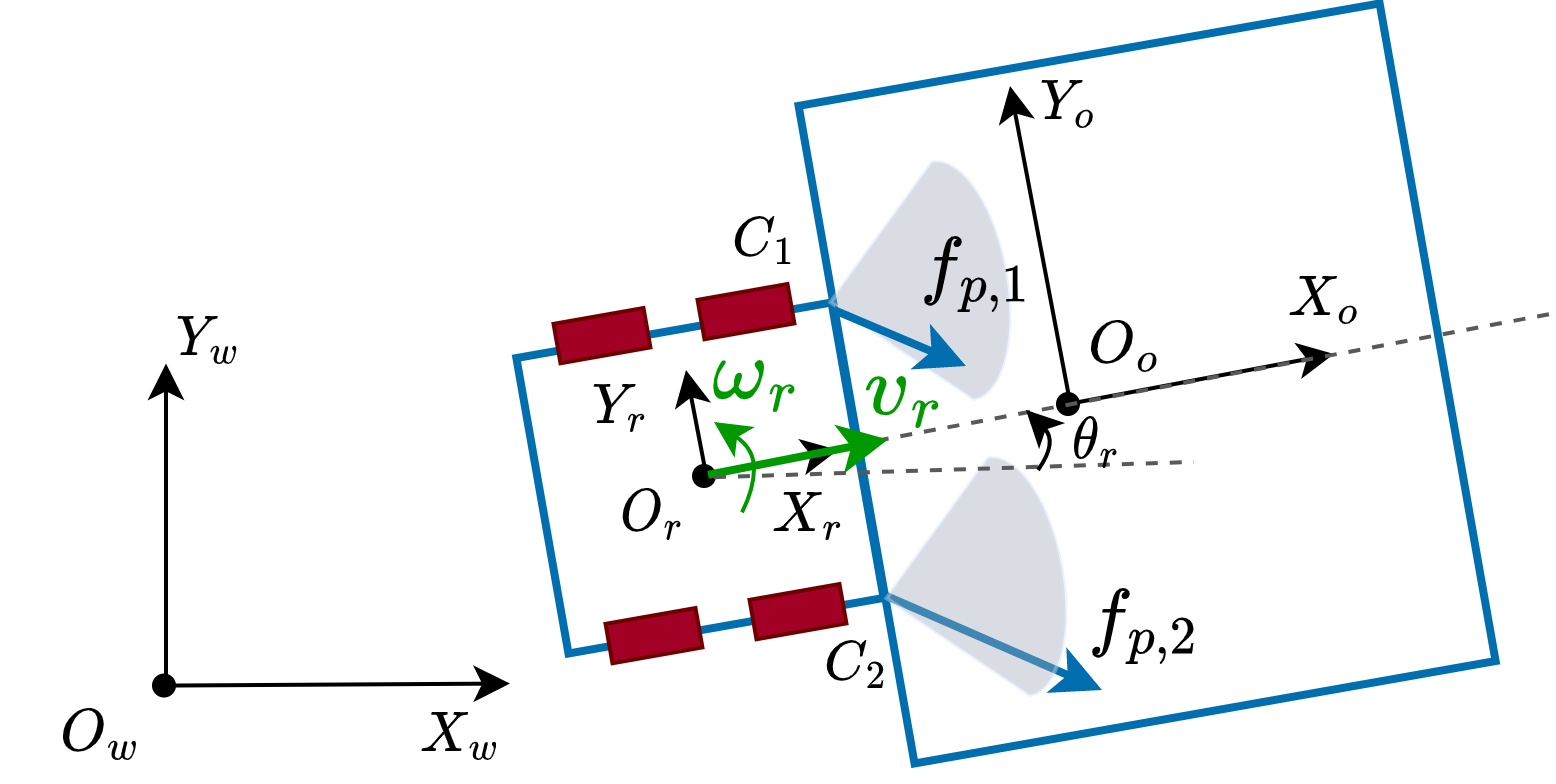}
		\par\end{centering}
	\caption{Schematic of the robot-object pushing system.\label{fig:frame}}
 \vspace{-0.6cm}
\end{figure}

As shown in Section.\ref{sec:graphical_derivation_single_point}, the maneuverability of the pushing system with a single-point contact is greatly restricted by using a nonholonomic rectangular mobile base.
We focus on pushing with line contact to improve maneuverability under stable pushing. Due to the complexity of directly imposing the friction cone constraint, we instead derive a simplified linear motion constraint tailored for the differential drive robot. This approach allows us to solve the stable pushing problem effectively. The Clearpath Husky and Boxer robot are used here, as shown in Fig. \ref{fig:push}.
The schematic of the pusher-slider system can be found in Fig. 
\ref{fig:frame}.

\subsection{Graphical derivation} \label{sec:graphical_derivation}
\begin{figure}[t]
\vspace{0.4cm}
	\begin{centering}
		\includegraphics[width=0.7\columnwidth]{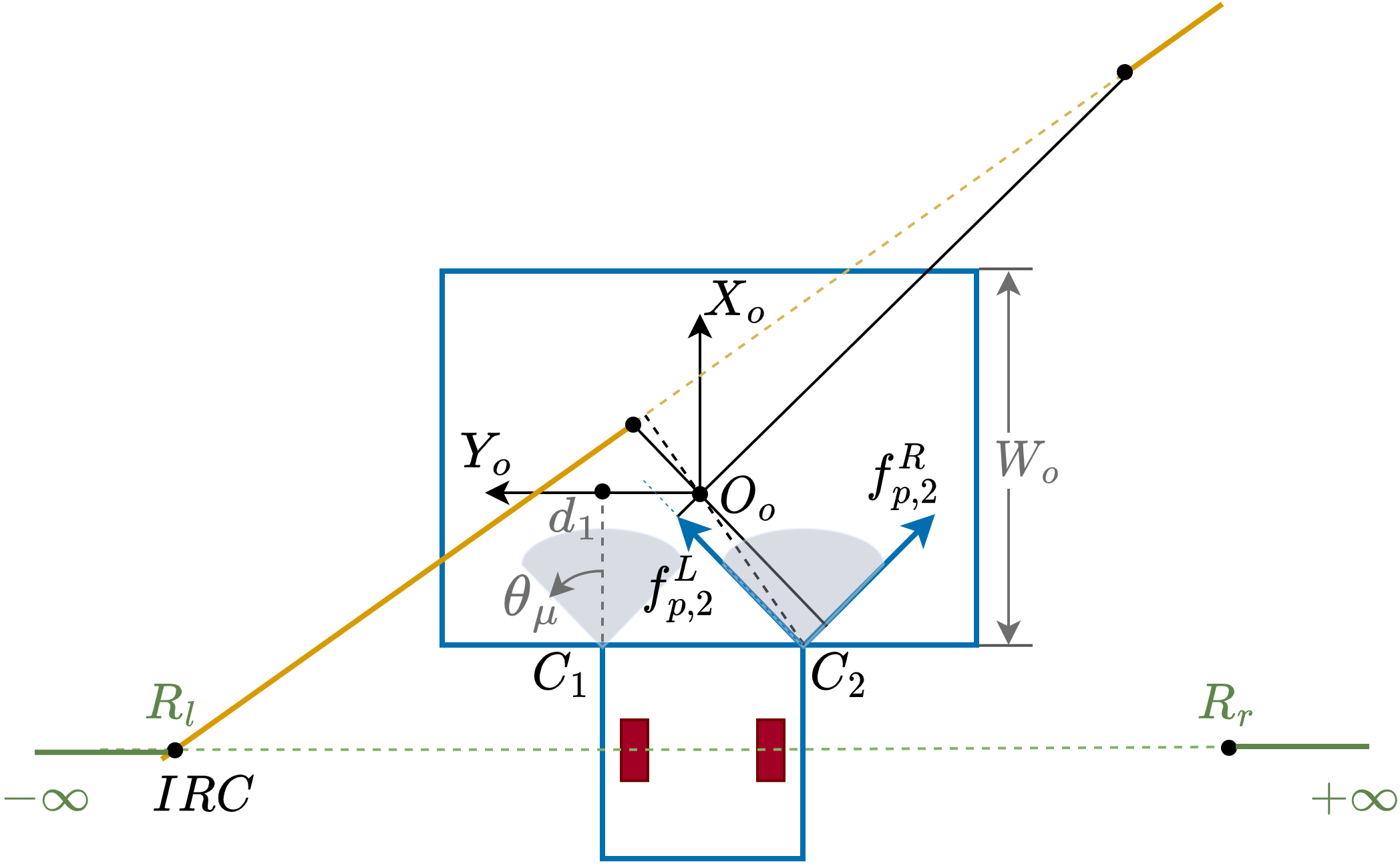}
		\par\end{centering}
	\caption{ Graphical demonstration for the sticking contact constraint. The intersection of the robot's and object's possible rotation center lies on the green lines. \label{fig:Graphical_derivation}
   }
\end{figure}

Building upon the derivation for point contact based on the graphical approach presented in Section.\ref{sec:graphical_derivation_single_point}, we extend it to the line contact case, as depicted in Fig. \ref{fig:Graphical_derivation}. The line contact can be simplified as two point contacts at the extreme points \cite{lynch1999locally}, ${}^{\cO} C_{\tn{i}} = [-W_{\tn{o}}/2, d_{\tn{i}}], i\in\{1,2\}$. The pushing force at contact points is denoted by 
$\vf_{\tn{p},i} = [f_{\tn{p},i}^{\tn{L}}, f_{\tn{p},i}^{\tn{R}}]^T \in \R^2$, including two components along the two edges of the friction cone. To ensure stiff contact between the robot and the object, the pushing forces, $\vf_{\tn{p},i}$, are limited within the friction cone.

A total generalized force, $\vf_\tn{p}=[f_{\tn{p}}^{\tn{L}}, f_{\tn{p}}^{\tn{R}}] \in \R^2$, and a corresponding generalized contact point, ${}^{\cO} C= [-W_{\tn{o}}/2, d], d\in[-\frac{L_o}{2},\frac{L_o}{2}]$, can be found, which are equivalent to the two pushing forces, $\vf_{\tn{p},i}, i\in{1,2}$, ensuring that the contact wrench exerted by the generalized force, ${}^\cO\vw_{\tn{p}}$, matches that of the pushing forces, ${}^\cO\vw_{\tn{p},1}$ and ${}^\cO\vw_{\tn{p},2}$: ${}^\cO\vw_{\tn{p}} = {}^\cO\vw_{\tn{p}, 1}+{}^\cO\vw_{\tn{p},2}$.

The generalized contact point shifts on the line segment $C_1C_2$, causing a tilt in the line of rotation centers $l$  (for details, please refer to
\cite{zhou2019pushing}). Consequently, this tilted $l$ intersects with the wheel axis of the robot, as illustrated in Fig. \ref{fig:Graphical_derivation}. Under the friction cone constraint, all the possible intersections form the line segment $[- \infty, R_\tn{l}]$ and $[R_\tn{r},+\infty]$. Obviously, the sticking constraint is transformed to a constrained motion set for the robot-object system.

\subsection{Algebraic derivation}
 
Now we derive the constrained motion set boundary using an algebraic approach.

The friction cone of the pushing force is
\begin{equation}
    \begin{aligned}
    \cF_{\tn{p},i} &= \{\vf_{\tn{p},i}\in\R^2~|~f_{\tn{p},i}^{\tn{L}} > 0, f_{\tn{p},i}^{\tn{R}} > 0\},~i=1,2.
    \end{aligned}
\end{equation}
Equivalently, the friction cone on $\vf_{\tn{p},i}$ can be written in a form of 
$    \begin{aligned}
    \vf_{\tn{p},i} = \lambda_{1,i} 
    \begin{bmatrix}
        1\\
        0
    \end{bmatrix}
     + \lambda_{2,i} 
     \begin{bmatrix}
         0\\
         1
     \end{bmatrix} ~|~ \lambda_{1,i}, \lambda_{2,i} >0
    \end{aligned}$
where $\lambda_{1,i}, \lambda_{2,i}$ are non-negative real numbers \cite{boyd2004convex}.
For each feasible friction force $\vf_{\tn{p},i} \in \cF_{\tn{p},i}$, it generates a wrench 
${}^\cO\vw_{\tn{p},i} = J_{\tn{p},i}\vf_{\tn{p},i}$ with $J_{\tn{p},i}$ the matrix that 
maps the contact friction force to a pusher wrench in the object's body frame.
\begin{equation}
    J_{\tn{p}, i} = 
    \begin{bmatrix}
        \tn{cos}(\theta_\mu) & \tn{cos}(\theta_\mu)  \\ 
        \tn{sin}(\theta_\mu) & -\tn{sin}(\theta_\mu) \\
        d_i \tn{cos}(\theta_\mu) + \frac{1}{2} W_{\tn{o}} \tn{sin}(\theta_\mu) &
        d_i \tn{cos}(\theta_\mu) - \frac{1}{2} W_{\tn{o}} \tn{sin}(\theta_\mu)
    \end{bmatrix}
\end{equation}

\begin{figure*}[t]
\vspace{0.4cm}
    \centering
	\includegraphics[width=1.6\columnwidth]{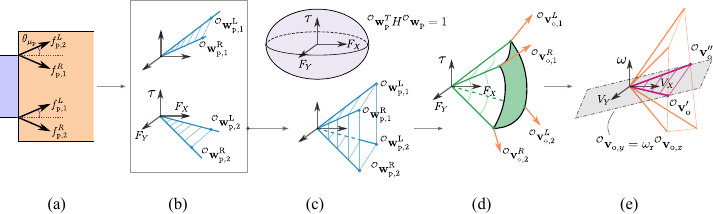}
	\caption{Illustration of the motion cone construction for planar pushing using a nonholonomic robot. (a) Friction cones. (b) Individual generalized friction cones. (c) Convex hull of the individual generalized friction cones (blue region) and the limit surface (light purple ellipsoid). (d) Feasible pusher wrenches (on the green surface) and force-motion model (orange vectors). (e) Motion cone of the object (area marked red).  \label{fig:motion_cone_construction}}
 \vspace{-7mm}
\end{figure*}

The friction cones on the contact points lead to the wrench cone. For each friction cone $\cF_{\tn{p},i}, i=1,2$, pusher wrenches ${}^\cO\vw_{\tn{p},i}^{\tn{L}}$ and ${}^\cO\vw_{\tn{p},i}^{\tn{R}}$ corresponding to the two-unit edges $\vf_{\tn{p},i}^{\tn{L}} = [0, 1]^\top$ and $\vf_{\tn{p},i}^{\tn{R}} = [1, 0]^\top$
, gives the edges of the wrench cones, as shown in Fig. \ref{fig:motion_cone_construction}b.
\begin{equation}\label{eq:friction_to_wrench}
    {}^{\cO}\cW_{\tn{p},i} = \{{}^\cO\vw_{\tn{p},i} = J_{\tn{p},i}\vf_{\tn{p},i}~|~ \vf_{\tn{p},i}\in\cF_{\tn{p},i}\}, ~i=1,2.
\end{equation}
where 
${}^\cO\vw_{\tn{p},i} = \lambda_{1,i} {}^\cO\vw_{\tn{p},i}^{\tn{L}} + \lambda_{2,i} {}^\cO\vw_{\tn{p},i}^{\tn{R}} 
                      = [
        \lambda_{1,i} \tn{cos}(\theta_\mu) + \lambda_{2,i} \tn{cos}(\theta_\mu), ~
        \lambda_{1,i} \tn{sin}(\theta_\mu) -\lambda_{2,i} \tn{sin}(\theta_\mu),~
        \lambda_{1,i} (d_i \tn{cos}(\theta_\mu) + \frac{1}{2} W_{\tn{o}} \tn{sin}(\theta_\mu)) +
        \lambda_{2,i} (d_i \tn{cos}(\theta_\mu) - \frac{1}{2} W_{\tn{o}} \tn{sin}(\theta_\mu)) ]^\top$.

Then the generalized wrench of the two pushing forces is 
\begin{equation}\label{eq:generalized_wrench}
    \begin{aligned}
        {}^{\cO}\vw_{\tn{p}} 
        & = \lambda_3{}^\cO\vw_{\tn{p,1}} + \lambda_4{}^\cO\vw_{\tn{p,2}}\\
        & = \lambda_3(\lambda_{1,1} {}^\cO\vw_{\tn{p},1}^{\tn{R}} + \lambda_{2,1} {}^\cO\vw_{\tn{p},1}^{\tn{L}}) \\
        &~~~ + \lambda_4(\lambda_{1,2} {}^\cO\vw_{\tn{p},2}^{\tn{R}} + \lambda_{2,2} {}^\cO\vw_{\tn{p},2}^{\tn{L}})
    \end{aligned}
\end{equation}
where $\lambda_j >0, ~j=3,4$.
Since $\lambda_{1,i}\lambda_j >0$, the feasible set of the generalized wrench in Eq. \eqref{eq:generalized_wrench} can be represented as a convex hull ${}^{\cO}\cW_{\tn{p}} $, as shown in Fig. \ref{fig:motion_cone_construction}c.
\begin{equation}\label{eq:GFC}
    {}^{\cO}\cW_{\tn{p}} = \tn{\tb{cvx\_hull}}(
        {}^\cO\vw_{\tn{p,1}}^{\tn{L}}, {}^\cO\vw_{\tn{p,1}}^{\tn{R}},
        {}^\cO\vw_{\tn{p,2}}^{\tn{L}}, {}^\cO\vw_{\tn{p,2}}^{\tn{R}}
    )
\end{equation}
As mentioned in Eq. \eqref{eq:force_motion_model}, the limit surface theory gives the mapping of the pushing force and the resulting object sliding motion. The direction of the object's twist is parallel to $H{}^\cO\vw_{\tn{p}}$. Combining with Eq. (\ref{eq:friction_to_wrench}), we can write all possible twists ${}^\cO\cV_{\tn{o}} = [{}^\cO v_{\tn{o},x}, {}^\cO v_{\tn{o},y}, {}^\cO \omega_{\tn{o}}]^\top $ of the object as:
\begin{equation}\label{eq:mc_first}
    {}^{\cO}\cV_{\tn{o}} = \{ k_{\tn{o}}H{}^\cO\vw_{\tn{p}}~|~{}^\cO\vw_{\tn{p}} \in {}^{\cO}{\cW}_{\tn{p}},~ k_{\tn{o}} \in \R^{+} \},
\end{equation}
where $k_{\tn{o}}$ is a magnitude parameter. 

For all pusher wrenches ${}^\cO\vw_{\tn{p}} \in {}^{\cO}{\cW}_{\tn{p}}$ that are on the ellipsoidal limit surface, the set of mapped object twists ${}^{\cO}\cV_{\tn{o}}$ is also a polyhedral cone since the mapping in Eq. (\ref{eq:mc_first}) is linear. Thus, we can compute the motion cone ${}^{\cO}{\cV}_{\tn{o}}$ by computing its edges, as shown in Fig. \ref{fig:motion_cone_construction}d. 

Additionally, since the object is pushed by the robot, which has a linear velocity $v_{\tn{r}}$ and 
angular velocity $\omega_{\tn{r}}$, without losing or sliding the contact, we have the object velocity 
\begin{equation}
    {}^{\cW} \vv_\tn{o} 
    = {}^{\cW} \vv_\tn{r} + {}^{\cW} R_{\cR} \cdot (\pmb{\omega_\tn{r}} \times {}^{\cR} \vx_\tn{o}  )_{(1:2)}
\label{eq:object_velocity_in_world}
\end{equation}
where $ {}^{\cR} \vx_\tn{o} =[d_{\tn{ro}}, {}^\cR y_{\tn{o}}, 0]^\top$ denotes the object position in the robot frame and $ \pmb{\omega_\tn{r}}=[0,0,\omega_\tn{r}]^\top$ corresponds to the pure rotation velocity vector of the robot. The subscript {(1:2)} indicates taking the first two dimensions of the vector.

After substituting Eq. \eqref{eq:robot_velocity_in_world} in Eq. \eqref{eq:object_velocity_in_world}, the velocity of the object expressed in the object frame can be achieved by multiplying ${}^{\cW} R_{\cR}^{-1}$ at both sides of Eq. \eqref{eq:object_velocity_in_world}, which yields:
\begin{equation}\label{eq:vwr}
    {}^\cO v_{\tn{o},x} = v_{\tn{r}}-\omega_{\tn{r}} {}^\cR y_{\tn{o}},~~
    {}^\cO v_{\tn{o},y} = \omega_{\tn{r}}d_{\tn{ro}},~~
    {}^\cO \omega_{\tn{o}} =\omega_{\tn{r}}.
\end{equation}

~It can be observed that Eq. (\ref{eq:vwr}) describes a plane ${}^{\cO}\cP_{\tn{o}}$ crossing the origin in the $\cO_{x-y-\omega}$ space:
${}^{\cO}\cP_{\tn{o}} = \{ {}^\cO\vv_{\tn{o}} ~|~{}^\cO v_{\tn{o},y} - d_{\tn{ro}}{}^\cO \omega_{\tn{o}} = 0 \}$, 
as shown in Fig. \ref{fig:motion_cone_construction}e. Combining Eq. (\ref{eq:mc_first})-(\ref{eq:vwr}), we can obtain the final possible twists of the object, known as the object motion cone, as the intersection of the set ${}^{\cO}\cV_{\tn{o}}$ and the plane ${}^{\cO}\cP_{\tn{o}}$:
$    {}^{\cO}\bar{\cV}_{\tn{o}} = \{ {}^\cO\vv_{\tn{o}} ~|~ {}^\cO\vv_{\tn{o}} \in {}^{\cO}\cV_{\tn{o}},~ {}^\cO\vv_{\tn{o}} \in {}^{\cO}\cP_{\tn{o}} \}$. 

The edges of the motion cone are computed as the intersection between the planes ${}^\cO \vv_{\tn{o},1}^R-{}^\cO \vv_{\tn{o},2}^R$, ${}^\cO \vv_{\tn{o},1}^L-{}^\cO \vv_{\tn{o},2}^L$ and the plane ${}^{\cO}\cP_{\tn{o}}$, which results in two edge vectors, ${}^\cO\vv_{\tn{o}}^{\prime}$ and ${}^\cO\vv_{\tn{o}}^{\prime\!\prime}$.
\begin{equation}\label{eq:}
    \begin{aligned}
        {}^\cO\vv_{\tn{o}}^{\prime} &= ({}^\cO \vv_{\tn{o},1}^L \times {}^\cO \vv_{\tn{o},2}^L) \times \vec{\vn}= 
        k_o \begin{bmatrix}
            -d_\tn{ro} \tn{cos}(\theta_\mu) \\
            -d_\tn{ro} \tn{sin}(\theta_\mu) \\
            -\tn{sin}(\theta_\mu)
        \end{bmatrix}
        \\
        {}^\cO\vv_{\tn{o}}^{\prime\!\prime} &= ({}^\cO \vv_{\tn{o},1}^R \times {}^\cO \vv_{\tn{o},2}^R) \times \vec{\vn} = 
        k_o \begin{bmatrix}
            -d_\tn{ro} \tn{cos}(\theta_\mu) \\
            d_\tn{ro} \tn{sin}(\theta_\mu) \\
            \tn{sin}(\theta_\mu)
        \end{bmatrix}
    \end{aligned}
\end{equation}
where $\vec{\vn}=[0,1,-d_{\tn{ro}}]$ is the normal vector to plane ${}^{\cO}\cP_{\tn{o}}$. 

The object motion cone can then be written as $   {}^{\cO}\bar{\cV}_{\tn{o}} = \lambda_5 {}^\cO\vv_{\tn{o}}^{\prime}  + \lambda_6 {}^\cO\vv_{\tn{o}}^{\prime\!\prime} ~|~ \lambda_5,\lambda_6 \in \R_{\geq0}$.
According to Eq. \eqref{eq:vwr}, we can achieve the corresponding motion cone for the robot, ${\cV}_{\tn{r}}$, with a linear mapping
$    \begin{bmatrix}
        v_\tn{r}\\ w_\tn{r}
    \end{bmatrix} = 
    \begin{bmatrix}
        1 & 0 & {}^\cR y_\tn{o} \\
        0 & 0 & 1
    \end{bmatrix}
    \begin{bmatrix}
        v_\tn{o,x} \\
        v_\tn{o,y} \\
        w_\tn{o}
    \end{bmatrix} |
    \begin{bmatrix}
        v_\tn{o,x} \\
        v_\tn{o,y} \\
        w_\tn{o}
    \end{bmatrix} \in {}^{\cO}\bar{\cV}_{\tn{o}}$.
Expressing the robot motion cone as a conical combination,
\begin{equation}
    \begin{aligned}
    \begin{bmatrix}
        v_\tn{r}\\ w_\tn{r}
    \end{bmatrix} =& \lambda_5 
    \begin{bmatrix}
        -{}^\cR y_{\tn{o}} \tn{sin}(\theta_{\mu}) -  d_\tn{ro} \tn{cos}(\theta_{\mu})\\
        - \tn{sin}(\theta_\mu)
    \end{bmatrix} + \\
    & \lambda_6
    \begin{bmatrix}
        {}^\cR y_{\tn{o}} \tn{sin}(\theta_{\mu}) -  d_\tn{ro} \tn{cos}(\theta_{\mu})\\
         \tn{sin}(\theta_\mu)
    \end{bmatrix}
    \end{aligned}
\end{equation}
from which we achieve the motion constraint on the robot input by finding the boundary of $w_\tn{r}/ v_\tn{r}$
\begin{equation}\label{eq:vw}
    \begin{aligned}
    k^{\prime\!\prime}v_{\tn{r}}^t \leq &\omega_{\tn{r}}^t \leq k^{\prime}v_{\tn{r}}^t 
    \end{aligned}
\end{equation}
where $v_\tn{r} \geq 0$, $k^{\prime\!\prime} = \frac{ \sin(\theta_\mu)}{ {}^\cR y_{\tn{o}} \sin(\theta_\mu) -  d_\tn{ro} \cos(\theta_{\mu})} $, $k^{\prime}=\frac{ \sin(\theta_\mu)}{ {}^\cR y_{\tn{o}} \sin(\theta_\mu) +  d_\tn{ro} \cos(\theta_{\mu})}$. 
It can also be regarded as a constraint on the curvature of the robot's trajectory, $k$. For simplification, we only plan for the pushes at the middle of the contact surface, where ${}^\cO y_{\tn{o}}=0$.

\section{Planning for Robot Pushing}

With the motion constraint derived in \eqref{eq:vw}, we now present a motion planner for robot pushing that keeps the object to be within its motion cone based on NMPC. 

\subsubsection{NMPC formulation}
We formulate a receding horizon optimization problem with $N$ time steps and planning horizon $N\Delta t$:
\begin{subequations}\label{eq:mpc_opti}
    \begin{align}
    \min\limits_{\vx_{\tn{r}}^{1:N}, \vu_{\tn{r}}^{0:N-1}} \quad          
                            & \sum_{\rebuttal{t}=0}^{N-1} J^t(\vx_{\tn{r}}^t, \vu_{\tn{r}}^t) + J^N(\vx_{\tn{r}}^N)\label{eq:cost_func} \\ 
    \text{s.t.}	\quad	    & \vx_{\tn{r}}^0 = \vx_{\tn{r}}(t_0), \quad \\
                            & \vx_{\tn{r}}^{t} = \vf_{\tn{r}}(\vx_{\tn{r}}^{t-1}, \vu_{\tn{r}}^{t-1}), \\
                            & \vh_{\tn{pushing}}(\vx_{\tn{r}}^t) \leq 0, \label{eq:push_constraints}\\ 
                            & \vh_{\tn{avoidance}}(\vx_{\tn{r}}^t) \leq 0, \label{eq:coll_constraints}\\ 
                            & \vu_{\tn{r}}^{t-1} \in \cU_{\tn{r}}, ~ \forall t\in\{1,\dots,N\},\nonumber
    \end{align}
\end{subequations}
where $\Delta t$ is the sampling time, $J^t$ denotes the cost term at stage $t$ and $J^N$ denotes the terminal cost, $\vx_{\tn{r}}(t_0)$ is the initial state of the robot, $\vf_{\tn{r}}$ is the robot dynamics model, $\cU_{\tn{r}}$ represents the robot's acceleration and angular acceleration limits. $ \vh_{\tn{pushing}}$ and $\vh_{\tn{avoidance}}$ respectively represent the path constraints for stable pushing and obstacle avoidance, which will be described in detail in the following. 

\subsubsection{Cost functions}
Let $\vp_{\tn{o}}^{\tn{g}}$ be the goal location that the object needs to be pushed to. We minimize the displacement between the object's terminal position with this goal. 
To this end, the terminal cost is defined as: 
    $J^N(\vx_{\tn{r}}^N) = q_{\tn{goal}}\norm{\vp_{\tn{o}}^N - \vp_{\tn{o}}^{\tn{g}}}$,
where the object's terminal position is $\vp_{\tn{o}}^N = \vp_{\tn{r}}^N + R(\theta_{\tn{r}}^N) [d_{\tn{ro}},0]^{\tn{T}}$ with $R(\cdot)$ the two-dimensional rotation matrix. $q_{\tn{goal}}$ is a tuning weight. 
The stage cost is to minimize the robot's linear and angular velocities to render it not to move too fast:$J^t(x_r^t, u_r^t) = q_{\tn{v}}(v_{\tn{r}}^t)^2 + q_{\omega}(\omega_{\tn{r}}^t)^2$, where $q_{\tn{v}}$ and $q_{\theta}$ are tuning weights. 

\subsubsection{Pushing constraints}
To make the robot keep contact with the object while pushing, the object's motion has to be within its motion cone at each time step. By combining the computed motion cone in Eq. \ref{eq:vw} with the continuous pushing constraint, the sticking contact constraints can be derived as follows: 
\begin{equation}\label{eq:pushing_const}
    \begin{aligned}
        &v_{\tn{r}}^t \geq 0, \\ 
        k^{\prime\!\prime}v_{\tn{r}}^t \leq &\omega_{\tn{r}}^t \leq k^{\prime}v_{\tn{r}}^t,
    \end{aligned}
\end{equation}

It indicates that the robot has to push forward the object, but its angular velocity should be within a motion cone related to the forward speed, which formulates the stable pushing constraints $\vh_{\tn{pushing}}$.

\subsubsection{Collision avoidance constraints}
\begin{figure}[t]
\vspace{0.5cm}
    \centering
	\includegraphics[width=0.7\columnwidth]{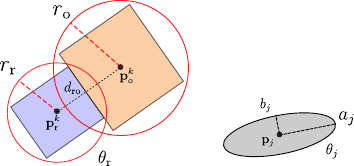}
	\caption{Illustration of collision avoidance between the robot-object system and the obstacle. \label{fig:collision_avoidance_ill}}
\end{figure}

For collision avoidance, we use two discs with radius $r = r_{\tn{r}}$ or $r_{\tn{o}}$ to circle the robot and object, respectively, as shown in Fig. \ref{fig:collision_avoidance_ill}. Each known obstacle $j = 1,\dots$ in the environment is modeled as an ellipse \cite{zhu2020icra} located at $\vp_j$ with semi-axis $(a_{j}, b_{j})$ and orientation $\theta_j$. Hence, the collision avoidance constraints $\vh_{\tn{avoidance}}$ are formulated as:
$	 (R(\theta_{j}) \vd_{j}^t)^{\tn{T}} 	
	 \begin{bmatrix}
		\frac{1}{(a_j+r)^2} & 0\\
		0 & \frac{1}{(b_j+r)^2}
	\end{bmatrix}
	R(\theta_{j}) \vd_{j}^t \geq 1$, 
where $\vd_{j}^t$ indicates the robot-obstacle relative position $\vp_{r}^t-\vp_j$, and the object-obstacle relative position $\vp_{\tn{o}}^t-\vp_j$ in which the object position is $\vp_{\tn{o}}^t = \vp_{\tn{r}}^t + R(\theta_{\tn{r}}^t)d_{\tn{ro}}$.

\section{Experimental Results}\label{sec:result}

\begin{figure*}[]
\vspace{0.5cm}
     \centering
     \begin{subfigure}[b]{0.22\textwidth}
         \centering
         \includegraphics[width=1.06\textwidth]{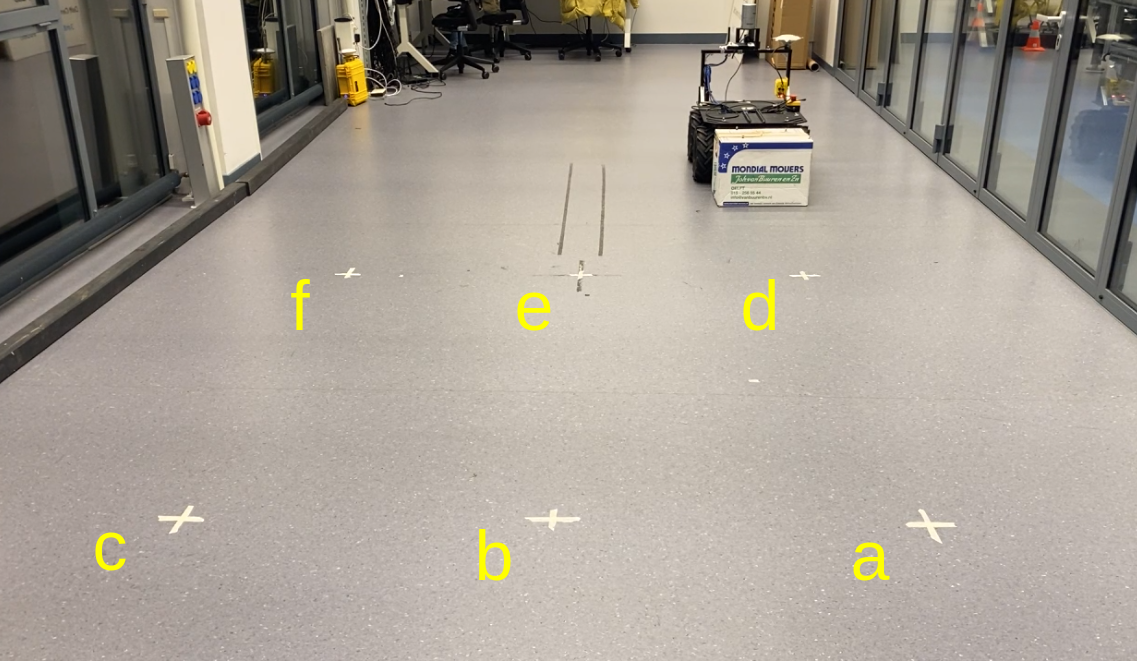}
         \caption{}
         \label{fig:multi_goal}
     \end{subfigure}
     ~~~
     \begin{subfigure}[b]{0.2\textwidth}
         \centering
         \includegraphics[width=\textwidth]{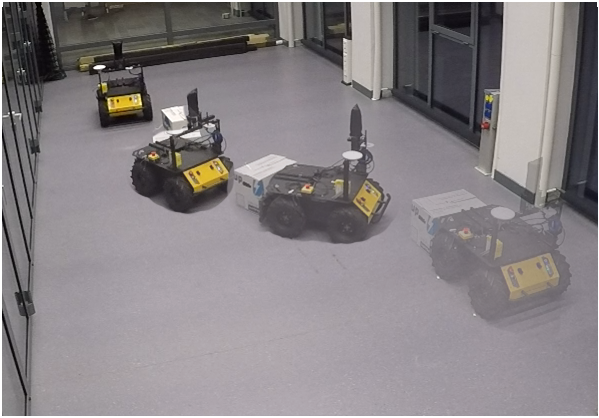}
         \caption{}
         \label{fig:husky_pushing_with_cons}
     \end{subfigure}
    ~
     \begin{subfigure}[b]{0.2\textwidth}
         \centering
         \includegraphics[width=\textwidth]{figs/result/real_exp/push_opacity.png}
        \caption{}
         \label{fig:husky_pushing_without_cons}
     \end{subfigure}
  \caption{(a) shows the selected pushing goals which are represented as white crosses on the floor. The corresponding goal-oriented pushing results are shown in Fig. \ref{fig:new_results} a-f. (b) and (c) separately show the experimental results of the robot pushing without and with the stiff contact constraint. The transparency of the robot and box in the image indicates their movement.}
  \label{fig:real_exp_2}
\end{figure*}

To validate the efficacy of our proposed method, we performed experiments using two robots, Clearpath Husky and Boxer, to test the stable pushing performance (Fig. \ref{fig:real_exp_2} and \ref{fig:real_exp_boxer}). Both the Husky and Boxer robots were differential-drive wheeled robots with rectangular shapes, respectively sized  $0.97 \times 0.67$ m and $0.75 \times 0.55$ m. Our experimental results demonstrated a 100\% stiff contact when applying the proposed concise stable pushing constraint. Additionally, we compared the proposed method with state-of-the-art pushing baselines to showcase the conciseness of our proposed constraint and the efficiency of stable pushing by effectively controlling object motion.

\subsection{Real-world Experiments using Husky and Boxer}

We carried out real-world experiments with two robots to demonstrate the efficacy of our proposed sticking contact constraint when stably pushing paper boxes. Our experiments utilized a motion capture system (OptiTrack) and a Kalman filter to collect information on robots, objects, and obstacles that operate at 120Hz. Control commands were calculated using our NMPC-based method on a laptop and sent to robots through WiFi and ROS, which operate at a frequency of 20Hz. We use the open source solver ACADOS \cite{Verschueren2021} to solve the NMPC problem, with a sampling time of $\Delta t = 0.1$ seconds, a planning horizon of $N = 20$ and tuning weights $q_\tn{goal}=1, q_\tn{v}=q_\omega=0.1$.

The Husky robot was equipped with a line bumper in the front, which acts as a pushing effector. It was used to push a large paper box measuring $0.32 \times 0.48 \times 0.48$ meters and weighing 2.8 kilograms. At the beginning of the push, the box was placed in contact with the robot center at a distance of $d_{\tn{ro}} = 0.66$ meters. The angle of the friction cone was set to $\theta_{\mu} = 12.00$ degrees. It is estimated by measuring the force which could pull the box at a constant speed, such that the pulling force is equal to the friction force: $F_\tn{pull} = F_{f} = \tan{\theta_\mu} \cdot m_\tn{o} g$. Then $\theta_\mu$ can be achieved as $\arctan(\frac{F_\tn{pull}}{m_\tn{o} g})$.

\begin{figure}
	\begin{centering}
		\includegraphics[width=1.\columnwidth]{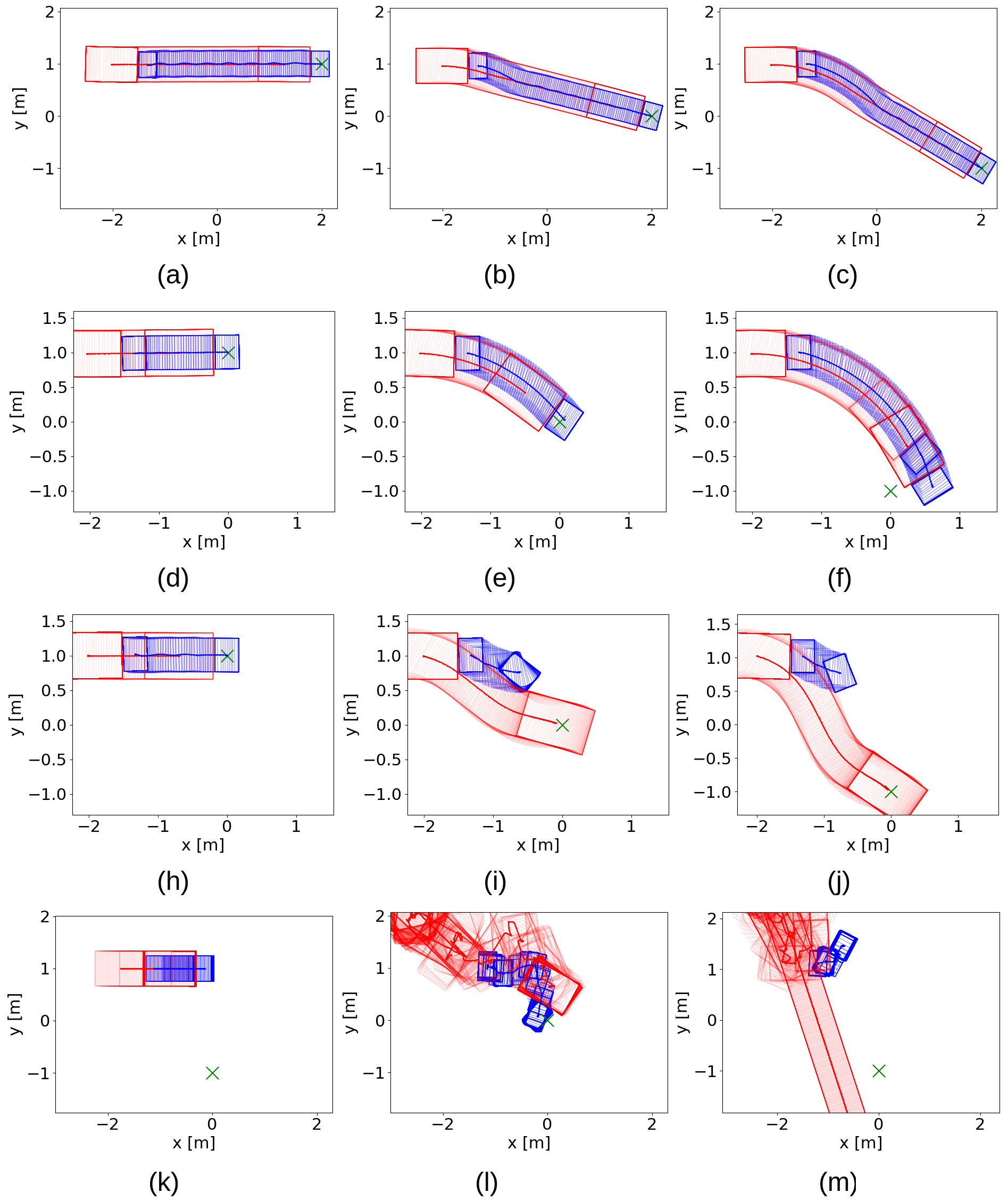}
		\par\end{centering}
 	\caption{
  (a-f) illustrate the stable pushing outcomes for the six chosen goals depicted in Fig. \ref{fig:multi_goal}. 
    For goals d, e, and f, 
    Fig. (h-j) additionally exhibit the pushing path without the sticking contact constraint, and Fig. (k-m) showcase the performance of the reactive pushing strategy.
  \label{fig:new_results}}	
    \vspace{-2.00mm}
\end{figure}

\begin{figure}
	\begin{centering}
		\includegraphics[width=0.6\columnwidth]{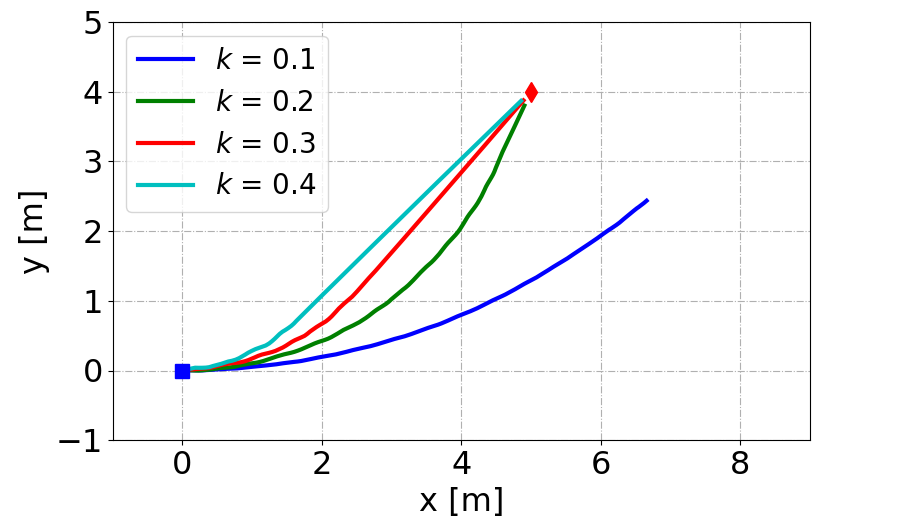}
		\par\end{centering}
 	\caption{Robot trajectories for pushing considering various limits of the robot trajectory curvature, where $k=k^{\prime}=\rebuttal{-}k^{\prime\prime}$. The blue square and the red diamond represent the start and the goal locations, respectively. The smaller the motion cone, the maneuverability of the robot is more limited.
  \label{fig:compare_mu}}	
\end{figure}

\begin{figure}
     \centering
     \begin{subfigure}[b]{0.35\textwidth}
         \centering
         \includegraphics[width=\textwidth]{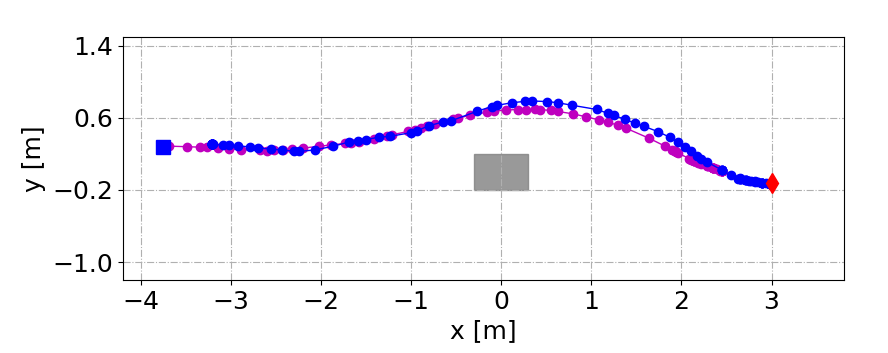}
         \label{fig:push_obs_avoid_traj}
     \end{subfigure}
    \caption{Experimental results of obstacle-aware robot pushing. The red and blue curves with dots represent the trajectories of the robot and the pushed object, respectively. The obstacle is marked in gray. \label{fig:push_obs_avoid}}
    \vspace{-2.00mm}
\end{figure}

Using the above setup, the limits of the robot trajectory curvature are calculated as $k^{\prime} = 0.32$ and $k^{\prime\prime} = -0.32$. Due to the size limitation of the motion capture system, we selected six pushing goals with coordinates (2,1), (2,0), (2,-1), (0,1), (0,0), and (0,-1) to evaluate the stable pushing performance, as shown in Fig. \ref{fig:multi_goal}. Starting from the initial position (-2,1), the Husky robot was tasked with pushing the paper box to the designated goal positions, as shown in Fig. \ref{fig:new_results} (a-f). The robot successfully maintained sticking contact with the object in all cases. Compared to trajectories without the stiff contact constraint (Fig.~\ref{fig:new_results} (h-j)), the object easily slides away while the robot moves (intuitive comparison can be found in Fig. \ref{fig:husky_pushing_with_cons} and \ref{fig:husky_pushing_without_cons}). However, the contact constraint also limited the maneuverability of the pushing system, so that the maximum curvature of the planned trajectory was bounded. Fig. \ref{fig:compare_mu} illustrates the relationship between maneuverability and motion cone. As a result, some pushing targets (e.g., Goal c in Fig. \ref{fig:new_results}) were unattainable within a limited time with the local NMPC planner. Reposition actions are required, so a global pushing planner will be the focus of our future research. Additionally, the proposed method can be easily extended to an obstacle-aware case, as shown in Fig. \ref{fig:push} and Fig. \ref{fig:push_obs_avoid}. A static obstacle is placed in front of the robot, and the object's goal location is behind it. The robot can successfully avoid the obstacle by maintaining both the stiff contact and obstacle avoidance constraints while pushing the object to the goal location.

\begin{figure*}
\vspace{0.5cm}
     \centering
     \begin{subfigure}[b]{0.25\textwidth}
         \centering
         \includegraphics[width=\textwidth]{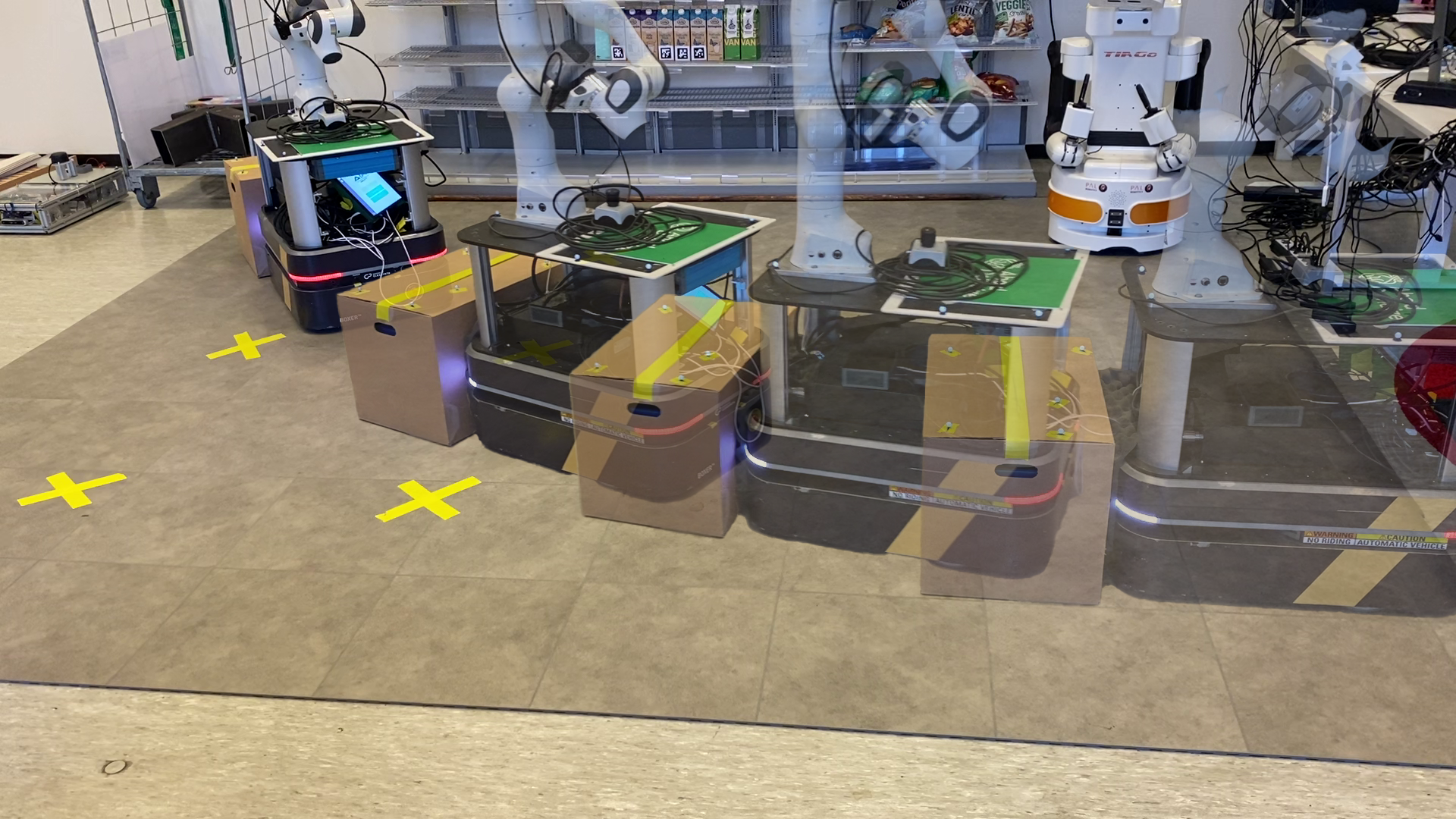}
         \caption{}
     \end{subfigure}
    ~
     \begin{subfigure}[b]{0.25\textwidth}
         \centering
         \includegraphics[width=\textwidth]{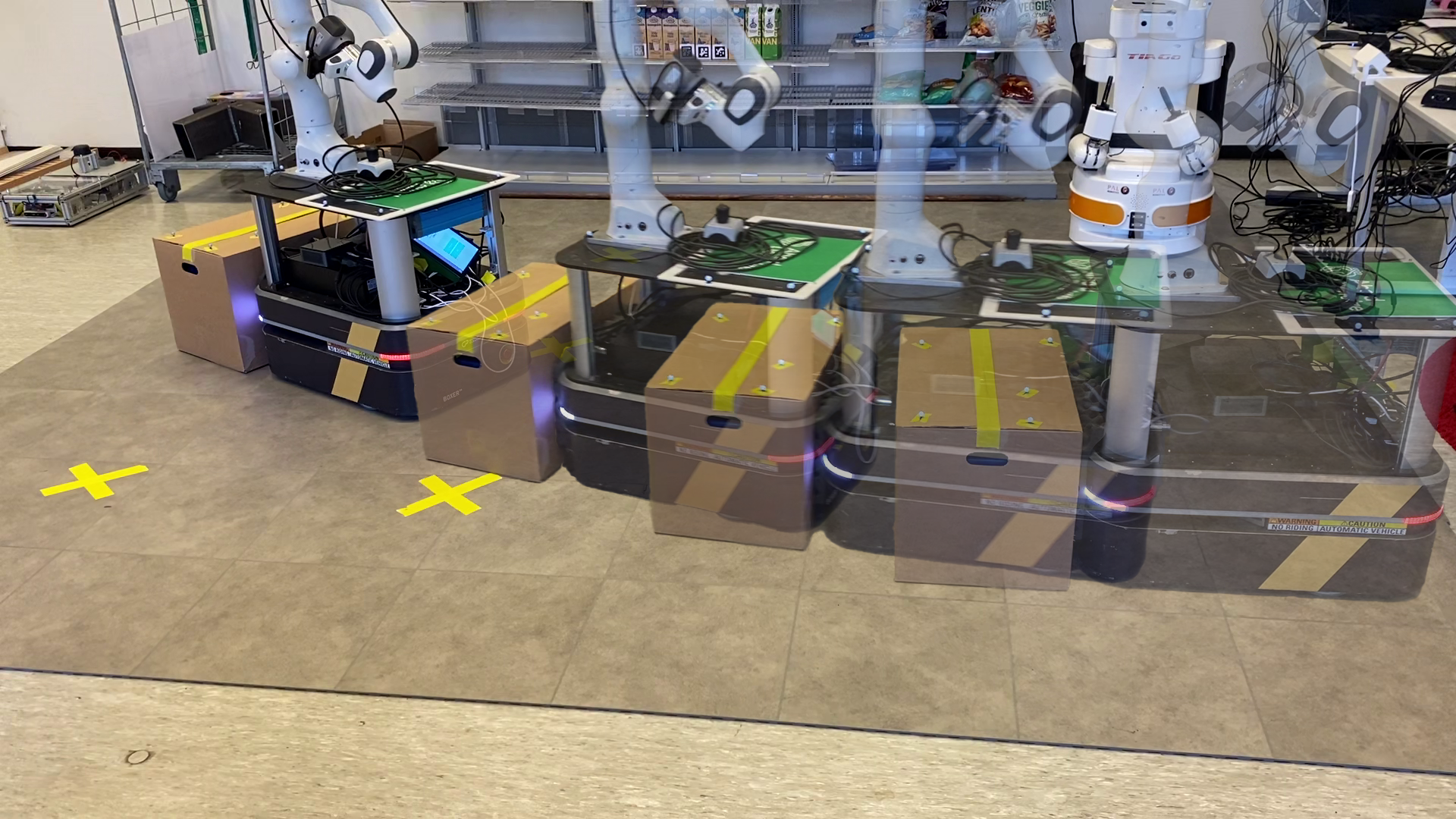}
         \caption{}
     \end{subfigure}
    ~
     \begin{subfigure}[b]{0.25\textwidth}
         \centering
         \includegraphics[width=\textwidth]{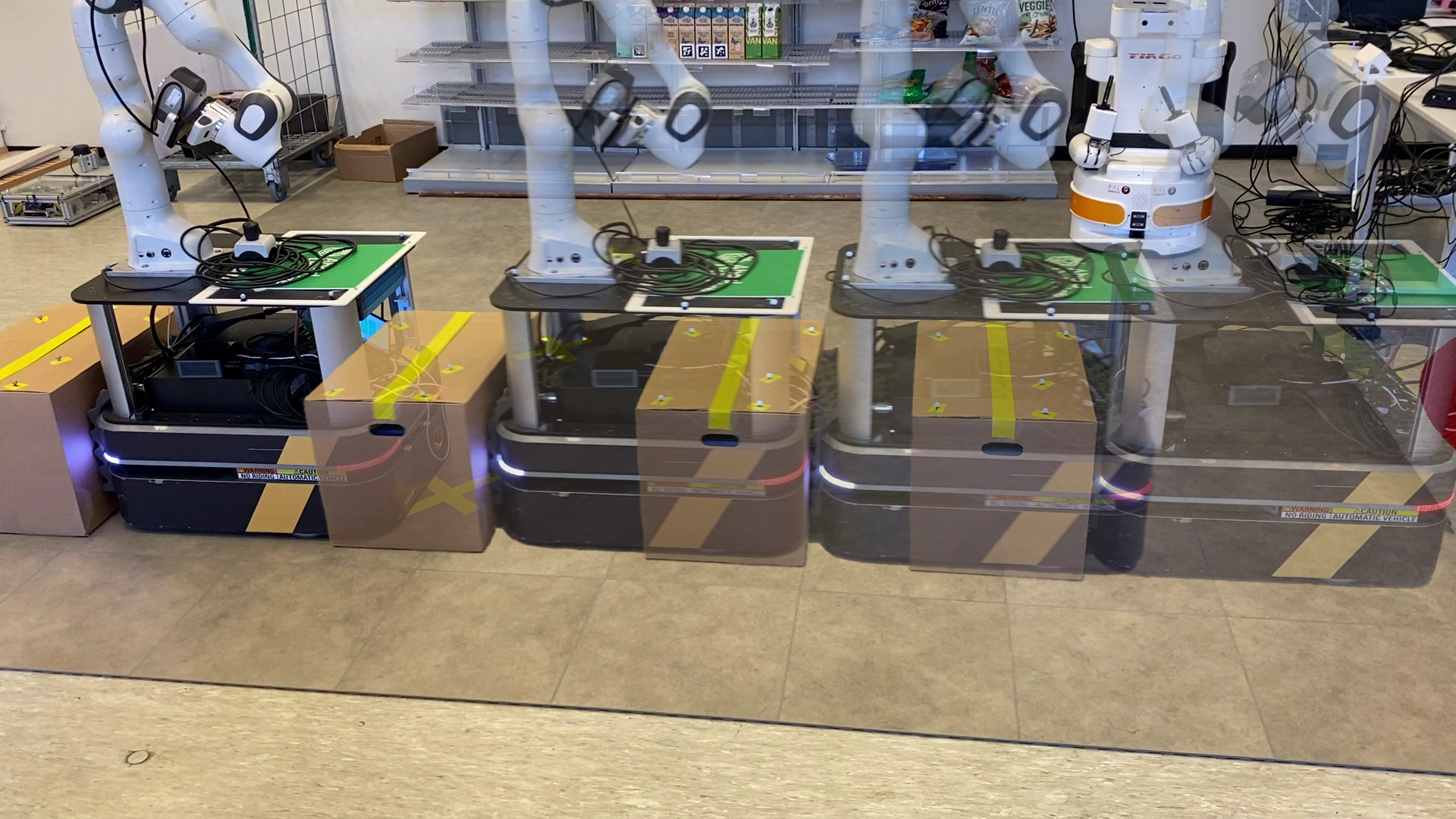}
        \caption{}
     \end{subfigure}
  \vspace{-2mm}
  \caption{Goal-targeted stable pushing with Boxer. Stiff contact is successfully maintained under the sticking contact constraint.}
  \label{fig:real_exp_boxer}
  \vspace{-5mm}
\end{figure*}

\begin{figure}
     \centering
     \begin{subfigure}[b]{0.32\textwidth}
         \centering
         \includegraphics[width=1\textwidth]{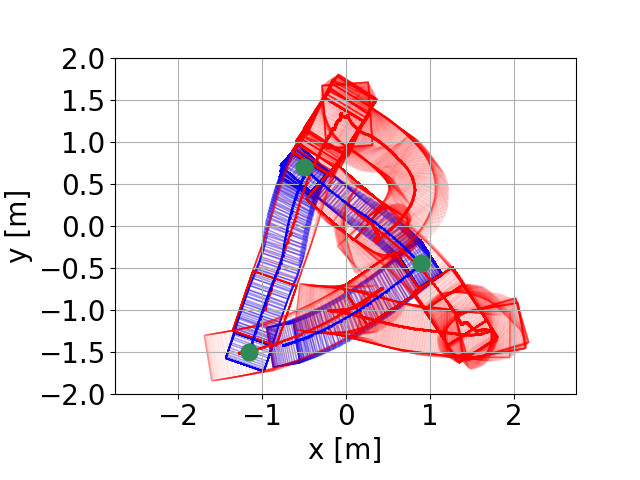}
         \caption{Push an object around the room.
         }
     \end{subfigure}
     \begin{subfigure}[b]{0.32\textwidth}
         \centering
         \includegraphics[width=1\textwidth]{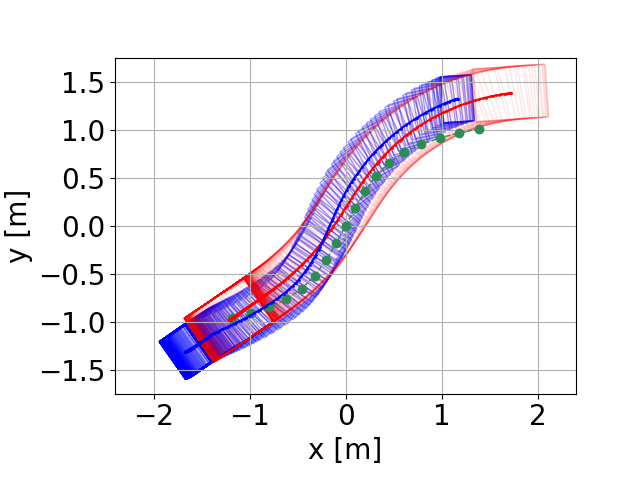}
         \caption{Path following while engaging in stable pushing.}
     \end{subfigure}     
  \caption{Stable pushing across different scenarios. The red and blue curves represent the trajectories of the robot and the pushed object, respectively. The reference waypoints are marked in green. In (b), A sponge sheet is sticked to the box to augment friction in the robot-object interaction, where $k^{\prime}=-k^{\prime\prime}=0.4$. For detailed information, we direct readers to our accompanying video.
  }
  \label{fig:different_scenarios}
\end{figure}

Furthermore, we aimed to comprehensively validate the effectiveness of our proposed stable pushing method under varying friction conditions using the Boxer robot within a distinct environment. A series of experiments were conducted to this end. In the initial phase, we conducted ablation studies to assess the effectiveness of the sticking contact constraint with box sized $0.39 \times 0.59$ m. Three pushing targets were selected, with five pushing trials conducted for each target. The outcomes of these ablation experiments are illustrated in Fig. \ref{fig:real_exp_boxer}, demonstrating an impressive 100\% success rate across all trials. Subsequently, we tried a new box sized $0.32 \times 0.48$ m and proceeded to an experiment where the robot pushed an object around the room. The implementation of the stiff contact constraint ensured that the robot maintained stiff contact with the object throughout the process. This strategic approach significantly reduced the need for frequent repositioning actions and requires only two designed switches. To further gauge the stability and robustness of our method, we designed a path tracking experiment. In this setup, the robot meticulously followed a predefined path while engaging in stable pushing. Both sets of experimental results are depicted in Fig. \ref{fig:different_scenarios} (shown in the attached video as well), illustrating the method's consistent performance across diverse scenarios. 

Overall, the outcomes of these comprehensive experiments demonstrate the robustness and efficacy of our proposed method across different friction conditions and robot platforms, underscoring its potential for real-world applications in robotics. 

\subsection{Comparison with the baseline approaches}\label{sec:expB}

What's more, to assess the performance of our proposed stable pushing method, we compared it to two existing baseline approaches, namely the reactive pushing strategy \cite{krivic2019pushing} and a Linear Time-Varying Model Predictive Control (LTV MPC) based stable pushing approach \cite{bertoncelli2020linear}. The comparison results are presented in Table \ref{tab:comparison_to_baselines}.

\begin{table}[h]
\resizebox{1\textwidth}{!}{\begin{minipage}{\textwidth}
\begin{tabular}{c|c|c|c|c}
\hline
 &
  \begin{tabular}[c]{@{}c@{}}Number of \\ hyper-\\ parameters\end{tabular} &
  \begin{tabular}[c]{@{}c@{}}Success rate\\ (For \\ Goal 1, 2, 3)\end{tabular} &
  \begin{tabular}[c]{@{}c@{}}Decision \\ variables\\ in MPC \\ (at time t)\end{tabular} &
  \begin{tabular}[c]{@{}c@{}}Solvable \\ with \\ commercial \\ solver\end{tabular} \\ \hline
\begin{tabular}[c]{@{}c@{}}Proposed\\ approach\end{tabular} & 1 & 100\%, 100\%, 0\% & 7  & Yes \\ \hline
\begin{tabular}[c]{@{}c@{}}Reactive\\ pushing\end{tabular}  & 5 & 100\%, 60\%, 0\%  & -  & -   \\ \hline
\begin{tabular}[c]{@{}c@{}}LTV\\ MPC\end{tabular}           & 5 & -                 & 11 & No  \\ \hline
\end{tabular}
\end{minipage} }
\caption{Comparison to the baselines}
\label{tab:comparison_to_baselines}
\end{table}

During the pushing process, the reactive pushing strategy attempts to minimize the angle between the object's movement direction and its direction toward the goal location. As a result, the robot must maneuver around the object to adjust its angle and sometimes reposition itself when the robot-object contact is lost. However, the core of the controller is a Proportional-Integral-Derivative (PID) controller, which is challenging to tune for optimal performance. Due to safety concerns, we tested this approach only in simulation. As shown in Fig. \ref{fig:new_results} (k-m), the robot often loses contact with the object, requiring time-consuming repositioning actions. Moreover, since the approach was originally designed for omnidirectional robots, it does not account for the motion constraints of nonholonomic robots. The robot sometimes bumps into the object while repositioning, adversely affecting pushing performance. In contrast, our proposed approach has demonstrated superior efficiency and pushing success rate for all three goals while maintaining a higher pushing success rate. The reactive pushing approach only achieves high success rates when the goal position is directly in front of the robot and is close to the initial position. To achieve the goals d, e, f, it has an average distance traveled by the robot and a time of $8.53$ m and $58.4$ s, respectively, while our proposed approach only takes $6.53$ m and $13.2$ s which saves $23.8\%$ and $77.4\%$ in these metrics. 

The LTV MPC-based pushing method shares the same motivation and mechanics as our proposed approach which is to add the friction cone constraint to guarantee stable pushing. However, the LTV MPC approach directly adds the stiff contact constraint to the optimization problem without any preprocessing. Consequently, it has four additional independent decision variables and four more hyperparameters to tune in the MPC formulation. We utilized the open source ACADOS solver to solve the MPC problem proposed in LTV MPC, which is unsolvable due to extra independent variables. Compared to other models, our concise stiff contact constraint requires only one hyperparameter ($k^{\prime}=-k^{\prime\prime}$) to tune and can be easily added to MPC-based navigation controllers.

\subsection{Sensitivity analysis}
Recognizing the inherent challenges in accurately measuring friction coefficients, we conducted a comprehensive sensitivity analysis. The primary goal was to determine the parameter $k$ without prior knowledge of the friction coefficient between the robot and the object. Additionally, we sought to comprehend how variations in the estimation of $k$ would impact the effectiveness of stable pushing. Subsequently, we assessed stable pushing performance for objects with distinct surface characteristics, including sponge sheet, foam sheet, and cardboard. Furthermore, recognizing the common occurrence of non-uniform mass distribution in unwieldy objects, we conducted experiments involving the rearrangement of the same set of objects within the box, thus achieving diverse mass distributions. This enabled us to investigate the method's robustness in scenarios where the assumption of uniform mass distribution is not perfectly upheld.

\begin{figure}
     \centering
     \begin{subfigure}[b]{0.5\textwidth}
         \centering
         \includegraphics[width=0.7\textwidth]{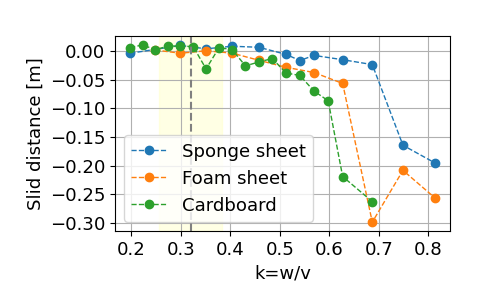}
         \caption{Slid distance for pushes along different curvatures, k, across different friction conditions which indicates the effectiveness of stable pushing.}
     \end{subfigure}\\
     \begin{subfigure}[b]{0.5\textwidth}
         \centering
         \includegraphics[width=0.7\textwidth]{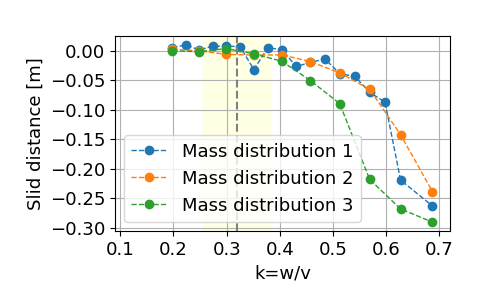}
         \caption{Slid distance for pushes along different curvatures, k, across different mass distributions which indicates the effectiveness of stable pushing.}
     \end{subfigure}
  \caption{Validate the effectiveness of stable pushing under different conditions. The grey dashed line corresponds to $k=0.32$. The yellow shading represents the range of $k\in [0.8k, 1.2k]$. Notably, the results illustrate that stable pushing—where the sliding distance is less than 0.05 m—can be realized whenever $k<0.32$, irrespective of alterations in friction conditions or mass distribution. Additionally, even when the hyperparameter in the stiff pushing constraint deviates by up to ±20\%, stable pushing remains intact.   
  }
  \label{fig:sensitivity_analysis}
\end{figure}

Because $k$ represents the limit of the robot trajectory curvature, our experimental setup entailed pushing various objects at a uniform speed of 0.1 m/s around a predetermined rotation center for a duration of 4 seconds. This rotation center, in turn, determines moving along a certain trajectory with curvature $k=w_\tn{r}/v_\tn{r}$. By measuring the displacement of the object's position in the robot frame at both the start and end of the trajectory, we quantified the cumulative slid distance of the object at different k. The outcomes of the experiments are depicted in Fig. \ref{fig:sensitivity_analysis}. Notably, when $k<k^{\prime}=0.32$ (for ${}^\cO y_{\tn{o}}=0$, where changing direction represents a symmetry case that we omit here), the object's slid distance remains at zero such that stable pushing is attainable. Conversely, when $k>k^{\prime}=0.32$, the assurance of stable pushing diminishes where the object slides. This observed trend persists across all tested friction conditions and mass distributions, underscoring the approach's capacity for generalization. Even when $k$ deviates by as much as ±20\%, the slid distance remains constrained to within 0.05 m.

\subsection{Discussion}
The proposed approach introduces a simple analytical stable pushing constraint, ensuring pushing stability under the line contact between the robot and the object. It is well-suited for objects with uniform mass distributions, and it can potentially be extended to handle cases with slightly nonuniform mass distributions and indeterminate anisotropic friction. Its simplicity is a notable feature, with only one hyperparameter requiring approximation. However, stable pushing imposes limitations on maximum trajectory curvature, which is decided by the friction condition between the robot-object interaction. Adding high friction coating will help to improve system maneuverability.

In contrast, there are widely-used learning-based pushing controllers utilize data-based pushing dynamics models, which do not consider the shape or mass distribution of the object \cite{mericcli2015push, arruda2017uncertainty, kopicki2017learning}. However, 
data-driven methods are known for their
data dependency, challenges in generalization, and susceptibility to Model Drift. Moreover, they neglect pushing stability, resulting in frequent object sliding and the need for time-consuming repositioning actions, especially problematic for nonholonomic mobile robots with limited maneuverability. 

The choice between stable pushing for regular-shaped objects and intermittent pushing for complex objects should be made based on the specific application's requirements and the characteristics of the objects involved.

\section{Conclusion}\label{sec:conclsuion}
This paper addresses the problem of using a differential-drive mobile robot to push an object to a goal location. We start by revisiting the pushing mechanics and highlighting the nonholonomic robot's challenges. To overcome the challenge, we propose a stable pushing approach that maintains a stiff line contact between the robot and the object, controlled by a stable pushing constraint. As a key contribution of this work, we provide an algorithm to simplify this constraint as a concise motion constraint for the robot. An NMPC-based planner is presented for stable pushing by considering the motion constraint. Our proposed method is more efficient than reactive pushing strategies, with a 23. 8\% reduction in the traveled trajectory length and a 77.4\% reduction in time.
Furthermore, our method is more concise than the LTV MPC-based stable pushing method, making it easier to implement. We validate our proposed method through real-world experiments with Husky and Boxer robots under different friction conditions. However, the stable pushing method has limitations in maneuverability. Our future research aims to design global policies that can further switch between contact surfaces to improve maneuverability.

\bibliographystyle{IEEEtran}
\balance
\bibliography{ref}

\end{document}